\documentclass[conference]{IEEEtran}
\IEEEoverridecommandlockouts
\usepackage{cite}
\usepackage{amsmath,amssymb,amsfonts}
\usepackage{algorithmic}
\usepackage{graphicx}
\usepackage{textcomp}
\usepackage{xcolor}
\usepackage{comment}
\usepackage{hyperref}
\usepackage{xcolor}
\usepackage[a4paper, total={184mm,239mm}]{geometry}
\usepackage[linesnumbered,ruled,vlined]{algorithm2e}
\def\BibTeX{{\rm B\kern-.05em{\sc i\kern-.025em b}\kern-.08em
    T\kern-.1667em\lower.7ex\hbox{E}\kern-.125emX}}
\renewcommand{\baselinestretch}{0.92}
\begin{document}

\title{BMPQ: Bit-Gradient Sensitivity-Driven Mixed-Precision Quantization of DNNs from Scratch\\
}

\author{\IEEEauthorblockN{Souvik Kundu, Shikai Wang, Qirui Sun, Peter A. Beerel, Massoud Pedram}
\thanks{$^{\dagger}$This work was supported in parts by NSF and DARPA with grant numbers 1763747 and HR00112190120, respectively.}
\IEEEauthorblockA{\textit{Electrical and Computer Engineering, }
\textit{University of Southern California, }
Los Angeles, USA \\
\{souvikku, shikaiwa, qiruisun, pabeerel, pedram\}@usc.edu}
}

\maketitle

\begin{abstract}
Large DNNs with mixed-precision quantization can achieve ultra-high compression while retaining high classification performance. However, because of the challenges in finding an accurate metric that can guide the optimization process, these methods either sacrifice significant performance compared to the 32-bit floating-point (FP-32) baseline or rely on a compute-expensive, iterative training policy that requires the availability of a pre-trained baseline. To address this issue, this paper presents BMPQ, a training method that uses \textit{\underline{b}it gradients} to analyze layer sensitivities and yield \textit{\underline{m}ixed-\underline{p}recision \underline{q}uantized} models. BMPQ requires a single training iteration but does not need a pre-trained baseline. It uses an integer linear program (ILP) to dynamically adjust the precision of layers during training, subject to a fixed hardware budget. 
To evaluate the efficacy of BMPQ, we conduct extensive experiments with VGG16 and ResNet18 on CIFAR-10, CIFAR-100, and Tiny-ImageNet datasets. Compared to the baseline FP-32 models, BMPQ can yield models that have $15.4\times$ fewer parameter bits with negligible drop in accuracy. 
Compared to the SOTA ``during training", mixed-precision training scheme, our models are $2.1\times$, $2.2\times$, and $2.9\times$ smaller, on CIFAR-10, CIFAR-100, and Tiny-ImageNet, respectively, with an improved accuracy of up to $14.54\%$.
\end{abstract}


\section{Introduction}
\label{sec:intro}
The prohibitively large computation and storage costs of current deep neural network (DNN) models pose a significant challenges in deployment of many large DNN models on resource-constrained IoT and edge devices. To address these issues, researchers have primarily focused on reducing the parameter budgets through various model compression techniques \cite{kundu2021dnr, deng2020model, kundu2021analyzing, kundu2020pre, wang2019haq, yang2021bsq}. In particular, $\textit{quantization}$ has proven to be a promising model compression scheme, that can essentially perform similar to the 32-bit floating-point (FP-32) parameter model and/or activation maps with low-precision, quantized, fixed-point values, requiring reduced storage and data-transfer costs than the FP-32 baselines. 

Early works of of binary neural networks (BNNs) \cite{hubara2016binarized} and XNOR-net \cite{rastegari2016xnor} with \textit{homogeneous-precision} (HPQ) models\footnote{All layer weights/activations have the same bit widths.}  showed the potential benefits of quantization. To address the issue of significant accuracy sacrifice of the HPQ models, more recent works have demonstrated the \textit{mixed-precision} quantization (MPQ) in which different layers can be assigned different bit widths based on the layer significance evaluated through various metrics, including Hessian spectrum \cite{dong2019hawq, yao2021hawq}.  

Most of the sensitivity-driven methods require
the presence of a baseline FP-32 pre-trained model.
Alternatively, quantization methods that rely on neural architecture search (NAS) \cite{wu2018mixed, wang2019haq}, to translate the model compression problem to a search problem of efficient bit-width assignment to different layers,  require a compute-intensive search procedure that is added on top of the training. Recently, researchers \cite{vasquez2021activation} have used intermediate activation densities to estimate the layer sensitivity and assign quantization bit widths, however, have not re-evaluated bit-width assignments, limiting performance. Moreover, their quantization method did not necessarily yield models satisfying a target hardware constraint.
Meanwhile, the increased demand for data-privacy has increased the need for on-device training and fine-tuning \cite{apple2017ondevice}. This trend makes many of these quantization-aware training methods impossible on resource-constrained devices. An alternative can be to rent costly GPU clusters, use quantization on private data, transfer the quantized model to resource-constrained device, and finally remove sensitive data from the server cluster before terminating the session. However, the prohibitively expensive training time may significantly increase the cluster cost,  that generally charges on hourly basis. This motivates the development of efficient training solutions that can yield mixed-precision quantized models with no baseline pre-training or iterative training.

\textbf{Our contribution.} We present an efficient ``during training" MPQ method that does not require a pre-trained model. To effectively search the large design space of MPQ, we decompose the core problem into two sub-problems. First, we use a novel \textit{bit-gradient-analysis} driven layer-sensitivity evaluation to rank the layers based on their significance. Second, using this information and a specific target hardware constraint, we formulate an integer linear program (ILP) to decide the bit-precision of each layer.
This combination of steps becomes the core of our
bit-gradient sensitivity driven MPQ method (BMPQ). BMPQ can be integrated into most training methods without any significant increase in training time because these extra steps are relatively low cost and are performed at regular intervals separated by multiple epochs of normal training. 

To demonstrate the efficacy of the proposed method, we perform extensive experimental evaluations on CIFAR-10, CIFAR-100, and Tiny-ImageNet with quantized VGG16 and ResNet18 models. Our experiments show that BMPQ can yield quantized models that require up to $15.4\times$ less storage compared to the FP-32 counterparts with comparable classification performance. Compared to the existing single-shot quantization approach \cite{vasquez2021activation}, our models can yield up to $2.9\times$ higher compression with an accuracy improvement of up to $14.54\%$.


\section{Related Work}
\label{sec:related}

\subsection{Quantization}
Quantization has improved the trade-off between accuracy and efficiency of NN models. Early works \cite{han2015deep} used rule-based strategies that required human expertise to quantize a model. 
Subsequent works focused on HPQ \cite{zhu2016trained, rastegari2016xnor} but often yielded reduced accuracy compared to the FP-32 baseline. To address this issue, works including non-uniform quantization \cite{zhang2018lq}, channel-wise quantization \cite{krishnamoorthi2018quantizing}, and progressive quantization-aware fine-tuning \cite{zhou2017incremental} were proposed.   

Recently, researchers have proposed using layer significance to guide quantization of different bit widths to different layers and thus boost the accuracy. However, the space of possible assignments is large.
In particular, for a model of $L$ layers with $N_B$ bit-width choices, there can be $(N_B)^{L}$ options to consider. To handle this issue, \cite{wang2019haq} converted the quantization problem into a reinforcement learning problem based on an actor-critic model. \cite{wu2018mixed} proposed a differentiable neural architecture search. 
Others use sensitivity analysis metrics \cite{dong2019hawq, yao2021hawq} to determine the layer importance and bit-width assignment but require a FP-32 pre-trained model or rely on iterative training \cite{yang2021bsq}.
Thus, despite all these efforts, a single-shot 
during-training MPQ approach that can yield similar to baseline performance while achieving ultra high compression has been largely missing.    

\subsection{PACT Non-Linearity Function for Activation}
The unbounded nature of the ReLU non-linear function can introduce significant approximation error when quantifying activations, particularly for low bit-precision activations  \cite{choi2019accurate}. Alternatively, clipped ReLU activations can provide bounded output, but finding a global clipping factor that maintains the model accuracy is quite challenging \cite{choi2019accurate}. A Parameterized Clipping Activation (PACT) function \cite{choi2018pact} with a per-layer parameterized clipping level $\alpha$ has been proven effective in the low-precision domain. For an input $a_{i}$, PACT non-linearity produces an output 
$a_{o}$ as follows:
\begin{align}
{a_{o}} &= 0.5(|a_{i}| - |a_{i}-\alpha| + \alpha) &=
    \begin{cases}
    0, &  a_{i} \in (-\infty, 0)\\
    a_{i}, & a_{i} \in [0, \alpha) \\
    \alpha, &  a_{i} \in [\alpha, +\infty)
    \end{cases}
\label{eq:pact}
\end{align}

The output is then linearly quantized to provide ${a_o}^q$. The computation of ${a_o}^q$ for a $k$-bit activation is 
\begin{align}
    {a_o}^q = \text{round}(a_{o}\cdot \frac{2^k - 1}{\alpha})\frac{\alpha}{2^k - 1}
\end{align}
The resemblance of PACT with ReLU increases with the increase of the value $\alpha$. Note that $\alpha$ is a trainable parameter and, during backpropagation, $\frac{\partial{{a_o}^q}}{\partial \alpha}$ is computed by using a straight-through estimator (STE) (see \cite{bengio2013estimating} for details). 

\section{Methodology}
\label{sec:method}
\subsection{Notation}
Let a $L$-layer DNN be denoted by $\textbf{Y} = f(\textbf{X}; \textbf{W})$, with each layer $l$ parameterized by $\textbf{W}_l$. 
A quantized version of $f$ is denoted as $f^Q(\textbf{X}^Q; \textbf{W}^Q)$ in which each layer $l$ is parameterized by $\textbf{W}^{q_l}_l$ with input activation tensor $\textbf{A}^{q_l}_l$. We train our model to minimize a loss $\mathcal{L}$ such that $f^Q$ can closely mimic $f$, and thus minimize any performance degradation. For HPQ, $q_l$ is fixed for any $l$, whereas for MPQ, $q_l$ may vary with $l$. 

\textbf{Definition 1. }\textit{Support bit widths}: For mixed-precision quantization, we define the support bit widths $S_q$ as the set of possible bit widths that can be assigned to the parameters of any layer $l$ of the model, excluding the first and last layers that are fixed to 16 bits as in  \cite{vasquez2021activation}. 

\subsection{Loss Bit Gradient}
The gradient of the loss with respect to a weight scalar $\frac{\partial \mathcal{L}}{\partial w}$ indicates the direction that reduces the output error at the highest rate. Moreover, larger magnitude gradients lead to more significant changes in the weights, and thus correlate well with larger weight significance \cite{dettmers2019sparse}. Inspired by this observation, we extend the notion of loss gradients to the bit-level for a quantized weight tensor, and propose a layer sensitivity metric that is driven by normalized bit gradients (NBG). 

For the $l^{th}$ layer of a DNN, the quantization 
of a floating-point tensor $\textbf{W}_l$ to a fixed-point (signed) tensor $\textbf{W}^{q_l}_l$ is

\begin{align}
    S_{w_l} = {\text{max}(|\textbf{W}_l|)}/({2^{q_l -1} - 1)}; \textbf{W}_l \in \mathbb{R}^{d_l}
\end{align}
\begin{align}
    \textbf{W}^{q_l}_l = \text{round}(\textbf{W}_l/S_{w_l})\cdot S_{w_l},
\end{align}

\noindent
where $d_l$ is the tensor dimension and $S_{w_l}$ is the quantization scaling factor. The quantized weights have a staircase function which is non-differentiable. To solve this problem we use STE, similar to other quantization methods \cite{yao2021hawq, dong2019hawq}. To reduce storage, it is recommended to store the fixed-point $(\textbf{W}^{q_l}_l/S_{w_l}) \in \{-2^{q_l -1},..., 2^{q_l -1}\}^{d_l}$ instead of $\textbf{W}^{q_l}_l$.   

As is typical, we convert the scaled quantized weights to their corresponding 2's complement representation \cite{rakin2019bit} as given by
\begin{align}
    {w}^{q_l}_l/S_{w_l} = -2^{q_l - 1}\cdot b_{q_l - 1} + \sum_{i=0}^{q_l - 2} 2^i\cdot b_i.
\end{align}
The loss bit gradient for each bit position is then derived as
\begin{align}
    \nabla_b\mathcal{L} &= [\frac{\partial \mathcal{L}}{\partial b_{q_l - 1}}, \frac{\partial \mathcal{L}}{\partial b_{q_l - 2}}, ... ,\frac{\partial \mathcal{L}}{\partial b_{0}} ] 
\end{align}
where
\begin{align}
\frac{\partial \mathcal{L}}{\partial b_{i}} &= \frac{\partial \mathcal{L}}{\partial {w^{q_l}_l}}\cdot  \frac{\partial {w^{q_l}_l}}{\partial b_{i}}.
\end{align}

For a layer $l$ with maximum support bit width $q_{max}$, we compute the loss bit gradient for each of the $q_{max}$ bits, yielding a $d_l \times q_{max}$ floating-point matrix. We sum the absolute values of each row to produce a $d_l \times 1$ vector. The layer NBG is set to be the average value of this vector.
Finally, we compute the epoch-normalized bit gradient (ENBG) of a layer as the mean of its NBGs over $i$ epochs. First-order derivatives like gradients can sometimes fail to capture the importance of a weight value based on its magnitude. However, the amortized nature of the ENBG computation over all weights and epochs makes the probability of this phenomenon occurring vanishingly small.

\textbf{Definition 2. }\textit{Epoch Intervals ($EI$):} We define the epoch intervals as the ranges of epochs over which we collect the NBG of each layer to calculate the ENBG. For training a model with periodic epoch intervals, the $k^{th}$ interval starts with the $(\sum_{i=0}^{k-1}ep^i_{int} +1)^{th}$ epoch where $ep^i_{int} = ep_{int}$ for any $i$. For aperiodic epoch intervals, we can use different $ep^i_{int}$ for different interval indices $i$. We use periodic $EI$ $ep_{int} = 20$. 

\subsection{ILP-Driven Iterative Bit-Width Assignment}
After each epoch interval, we re-assign the bit widths to maximize performance by formulating an ILP that maximizes the total layer sensitivity subject to a given hardware constraint $C$.  In particular, to capture the bit-width assignment at the start of the $k^{th}$ interval for a layer $l$, we introduce a constrained integer variable ${\Omega^k}_l$, which is multiplied by the negative ENBG of that layer $g^{k-1}_1$. Next we solve the following problem:

\begin{align}
    \text{Objective:} \text{   minimize} \sum_{l=0}^{L-1}({{g}^{k-1}_l.\Omega^k}_l),\\
    \text{Subject to:} \sum_{l=0}^{L-1}\Phi(\Psi\{{\Omega^k}_l\}) \le C
\end{align}
\noindent
Here, $\Psi(.)$ is a function that translates the assignment variable ${\Omega^k}_l$ to the corresponding bit width $q^k_l$ and $\Phi(.)$ is a function that translates the bit width to a cost associated with the layer.
For example, if $C$ is a memory-constraint, then $\Phi(.)$ converts the bit-width assignment to a measure of memory usage.
Notice that, in reality ${g}^{k-1}_l$ is computed assuming the bit-width assignments of other layers are fixed despite the fact that they may change after the ILP.
This approximation enables our ILP optimization to search over a large combinatorial space and any error associated with the approximation is mitigated by the repeated re-evaluation of the ILP after each epoch interval.
 

Recently, an iterative training approach \cite{yao2021hawq} has also used an ILP to assign layer bit widths but only as a post-training optimization.\footnote{We do not quantitatively compare our work with this work because this work uses an iterative (rather than a single-shot) training approach and thus it does not constitute an ``apple-to-apple" comparison to our work.} Moreover, reference \cite{yao2021hawq} computed a layer-variable coefficient based on an $L_2$-norm of the difference between FP-32 and quantized weights of that layer, which may further increase the storage overhead. In contrast, we use the negative ENBG values as coefficients of the respective  ${\Omega^k}_l$ that does not require any $L_2$-difference compute/storage cost.
\begin{figure}[!t]
\includegraphics[width=0.63\linewidth]{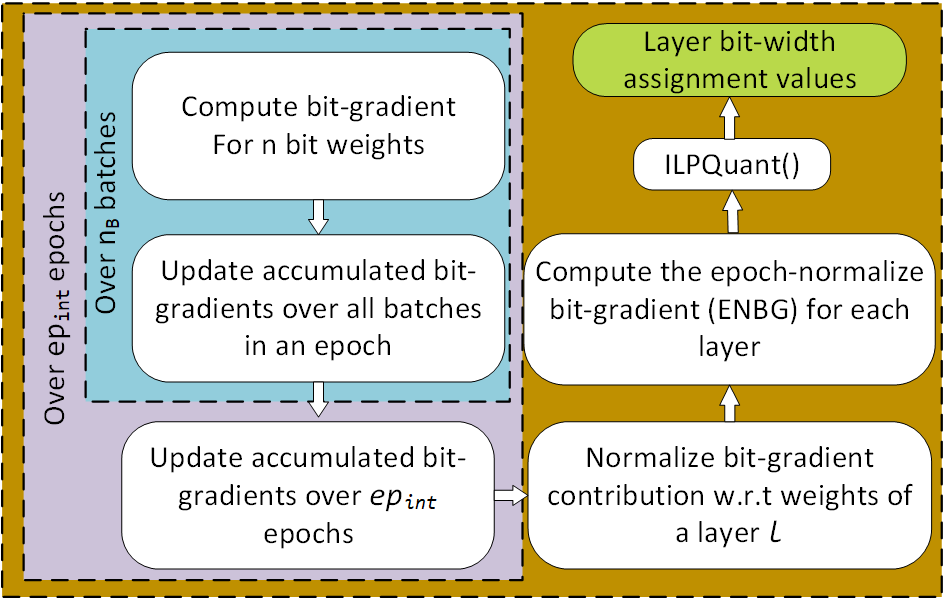}
\centering
\vspace{-0.2cm}
\caption{Step-wise description of layer bit-width evaluation.}
\label{fig:hybridizewtquant}
\vspace{-4mm}
\end{figure}
\subsection{BMPQ Training}
For the initial $ep_w$ warm-up epochs, we train the model with each layer quantized to $max$($N_1$, $N_2$, \ldots, $N_m$) bits where $S_q = [N_1,  \ldots, N_m]$. Throughout the training, we follow the recommendation of \cite{vasquez2021activation} to quantize the activation of layer $l$ with the same number of bits as that used for the weights of that layer. During the weight update, we first allow the weights to be the updated FP-32 values, and then quantize to a specific bit-precision and compute the loss based on forward-pass evaluation with the quantized weights/activations.  We use the ReLU non-linearity for the last layer and use the PACT nonlinearity for all other intermediate layers with low bit-precision. For a bit width $N_i = 2$, we use ternary weight quantization \cite{li2016ternary} to minimize the Euclidean distance between the FP-32 and quantized weights. The fact that we perform quantized training for $ep_{int}$ epochs after every bit-width assignment iteration helps us avoid post-training fine-tuning operations to maintain accuracy. The step-wise details of the $\text{evalENBG}(.)$ function is presented in Fig. \ref{fig:hybridizewtquant}. 
\section{Experiments}
\label{sec:expt}
\subsection{Experimental Setup}
\textbf{Models and Datasets.}
 We selected three widely used datasets, CIFAR-10, CIFAR-100, and Tiny-ImageNet and chose popular CNN models for image classification, VGG16 \cite{simonyan2014very} and ResNet18 \cite{he2016deep}. 
 For all the datasets we used standard data augmentations (horizontal flip and random crop with reflective padding) to train the models with  a batch size of 128.
 
 \textbf{BMPQ training settings.} For CIFAR-10 and CIFAR-100, we trained our models for 200 epochs with initial learning rate (LR) of 0.1 that decayed by 0.1 after 80 and 140 epochs. For Tiny-ImageNet, we used 100 training epochs and used decay epochs as 40 and 70 keeping other hyperparameters the same as that for CIFAR. We used $S_q$ of 4 and 2 bits for all our training but fixed the first and last layers to have 16-bits. For the ResNet models, we ensured the downsampling layers have the same bit-width assignment as its input layer \cite{vasquez2021activation}.  

\subsection{Results}
%
%
Table \ref{tab:results_bmpq} shows the performance of the BMPQ trained models compared to their respective full-precision counterparts. In particular, for CIFAR-10 dataset the BMPQ trained models can yield a model compression ratio\footnote{We define compression ratio as the ratio of bits required to store a FP-32 model to that required by the BMPQ model of same architecture.} of up to $15.4\times$ while sacrificing a drop in  absolute accuracy of up to only $0.70\%$. Similarly, for CIFAR-100 and Tiny-ImageNet, the BMPQ generated models provide near full precision performances while yielding a compression ratio of up to $15.4\times$ and $10\times$, respectively. This clearly shows the efficacy of BMPQ generated models to retain baseline performance while requiring  lower model storage cost.

\begin{figure}[!t]
\includegraphics[width=0.9\linewidth]{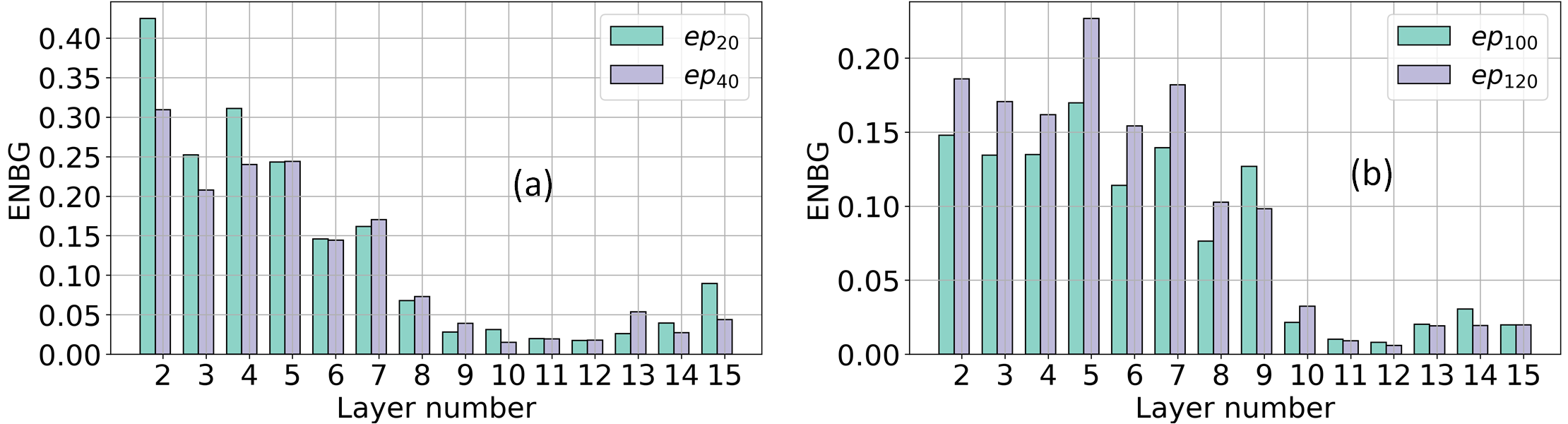}
\centering
\vspace{-0.2cm}
\caption{(a), (b) Layer sensitivities based on ENBG for VGG16 on CIFAR10. $ep_i$ indicates that the normalization was performed after $i^{th}$ epoch.}
\label{fig:ENBG_sensi}
\vspace{-2mm}
\end{figure}

\textbf{Analysis of ENBG at various iterations.} We analyzed the ENBG snapshots of VGG16 (on CIFAR-10) at the end of different epoch values. In particular, we chose two early-stage training epochs 20 and 40 and two mid-level training epochs 100 and 120. As shown in Fig. \ref{fig:ENBG_sensi}(a), the ENBG represented layer sensitivity  changes significantly between epoch 20 and 40, hinting at the need for iterative re-evaluation during training. Also, a similar trend is observed at the mid-level training epochs (Fig. \ref{fig:ENBG_sensi}(b)), that forces the ILP to re-assign the $10^{th}$ and $14^{th}$ layer from 2-b to 4-b and 4-b to 2-b, respectively. 

\begin{table}
\begin{center}
\tiny\addtolength{\tabcolsep}{-4.5pt}
\begin{tabular}{|c|c|c|c|c|}
\hline
Dataset & Model & layer-wise bit width & Test Acc ($\%$) & Compression\\
{} &  {} & {}   & {} & ratio ($r^{32}_M$)\\
\hline
\hline
{} & {} & Full precision & 93.9 & $1 \times$ \\
{} & {VGG16} & [16, 4, 4, 4, 4, 4, 4, 4, 4, 4, 2, 2, 2, 2, 4, 16] & 93.56 & $10.5 \times$ \\
{CIFAR-10} & {} & [16, 4, 2, 4, 4, 2, 2, 2, 2, 2, 2, 2, 2, 2, 2, 16] & 93.21 & $15.4 \times$\\
\cline{2-5}
{} & {} & Full precision & 95.14 & $1 \times$\\
{} & {ResNet18} & [16, 2, 2, 4, 2, 4, 4, 2, 2, 4, 4, 4, 2, 2, 2, 2, 2, 16] & 94.54 & $13.4 \times$\\
\hline
\hline
{} & {} & Full precision & 73 & $1 \times$\\
{} & {} &  [16, 4, 4, 4, 4, 2, 4, 2, 2, 2, 2, 2, 2, 2, 4, 16] & 72.2 & $14.6 \times$\\
{CIFAR-100} & {} & [16, 4, 2, 4, 4, 2, 2, 2, 2, 2, 2, 2, 2, 2, 2, 16] & 71.26 & $15.4 \times$\\
\cline{2-5}
{} & {} & Full precision & 77.5 & $1 \times$\\
{} & {ResNet18} & [16, 2, 2, 4, 2, 4, 4, 4, 2, 4, 4, 2, 4, 4, 4, 4, 2, 16] & 75.98 & $9.4 \times$\\
\hline
\hline
{} & {VGG16} & Full precision & 60.82 & $1 \times$\\
{Tiny-ImageNet} & {} & [16, 4, 4, 4, 4, 4, 4, 2, 4, 4, 2, 2, 4, 2, 4, 16] & 59.29 & $10 \times$\\
\cline{2-5}
{} & {ResNet18} & Full precision & 64.15 & 1 $\times$\\
{} & {} & [16, 2, 2, 2, 2, 2, 2, 2, 2, 2, 2, 2, 4, 4, 4, 4, 4, 16] & 63.27 & $8.8 \times$\\
\hline
\end{tabular}
\end{center}
\caption{Performance of BMPQ generated models compared to the respective baseline full precision (FP-32) models.}
\vspace{-8mm}
\label{tab:results_bmpq}
\end{table}
\subsection{Comparison with Single-Shot Training}

Table \ref{tab:compare_sota} presents the comparison of the presented BMPQ with the 
recently proposed single-shot MPQ method  \cite{vasquez2021activation} on CIFAR-10\footnote{The original paper used VGG19 compared to VGG16 of ours.}, CIFAR-100, and Tiny-ImageNet. In particular, we can see that BMPQ models provide an improved accuracy of up to $14.54\%$ with up to $2.9\times$ less parameter bits. Note, considering the training epochs of \cite{vasquez2021activation}, to have a fair comparison we report the accuracy of our models after 120, 120, and 60 epochs for CIFAR-10, CIFAR-100, and Tiny-ImageNet, respectively.

\begin{table}[!h]
\begin{center}
\tiny\addtolength{\tabcolsep}{-5.0pt}
\begin{tabular}{|c|c|c|c|c|}
\hline
 Model  & Dataset & AD \cite{vasquez2021activation} & BMPQ & Improved \\
  {}  & {} & Acc ($\%$) & Acc ($\%$)  & compression\\
\hline
VGG16 & CIFAR-10 & 91.62 &  \textbf{92.28} & $2.1\times$ \\
\hline
ResNet18 & CIFAR-100 & 71.51 &  \textbf{73.96} & $2.2\times$ \\
\hline
ResNet18 & Tiny-ImageNet & 44 &  \textbf{58.54} & $2.9\times$ \\
\hline
\end{tabular}
\end{center}
\caption{Comparison with single-shot MPQ achieved through analysis of activation density (AD).}
\vspace{-8mm}
\label{tab:compare_sota}
\end{table}
%

\subsection{Discussion}

\textbf{Memory saving for inference.}
Let layer $l$ of an $L$-layer model have $p_l$ parameters. Represented using homogeneous FP-32, the total storage requirement (MB) of the model weights can be given by
\begin{align}
    M_{\textit{fp32}} = 4*(\sum_{i=0}^{L-1}\frac{p_l}{2^{20}}). 
\end{align}
For a BMPQ generated model, the weight storage cost (MB) and corresponding compression ratio $r^{32}_M$ (and $r^{16}_M$ compared to a 16-b baseline) can be computed as
\begin{align}
    M_{\textit{BMPQ}} &= (\frac{4}{32})*(\sum_{i=0}^{L-1}\frac{p_l \cdot q_l}{2^{20}}) \\
    r^{32}_M &= \frac{M_{\textit{fp32}}}{M_{\textit{BMPQ}}} \text{ and } r^{16}_M = 0.5*r^{32}_M. 
\end{align}
Note that, for each layer, we need to store only one scaling factor in FP-32. Hence,
its overhead is negligible and ignored.
Column $5$ in Table \ref{tab:results_bmpq} shows the corresponding $r^{32}_M$ for our BMPQ models.
\section{Conclusions}
\label{sec:conc}
This paper presented a MPQ training method driven by epoch-normalized bit-gradients  without the requirement of any pre-training. Our proposed ENBGs capture the sensitivity of DNN layers and drive an ILP formulation that iteratively assigns MPQ bit widths after certain epochs during training. Our results demonstrated the efficacy of BMPQ models when compared to both FP-32 baselines and existing single-shot MPQ schemes. With the growing demand of privacy-preserving, on-device training and inference, we believe this work will act as a foundation for energy-efficient, on-device quantization.  

\bibliographystyle{IEEEtran}
\bibliography{IEEEabrv,biblio}

\end{document}